\documentclass{article}

\usepackage[nonatbib, final]{nips_2016}

\usepackage{wrapfig,graphicx}

\usepackage[utf8]{inputenc} % allow utf-8 input
\usepackage[T1]{fontenc}    % use 8-bit T1 fonts
\usepackage{hyperref}       % hyperlinks
\usepackage{url}            % simple URL typesetting
\usepackage{booktabs}       % professional-quality tables
\usepackage{amsfonts}       % blackboard math symbols
\usepackage{nicefrac}       % compact symbols for 1/2, etc.
\usepackage{microtype,amsmath}      % microtypography
\usepackage[algoruled,titlenumbered,noend]{algorithm2e}
\usepackage{subcaption}
\newtheorem{THEOREM}{Theorem}
\newenvironment{theorem}{\begin{THEOREM} \hspace{-.85em} {\bf :} }%
                        {\end{THEOREM}}

\newtheorem{LEMMA}[THEOREM]{Lemma}
\newenvironment{lemma}{\begin{LEMMA} \hspace{-.85em} {\bf :} }%
                      {\end{LEMMA}}
\newtheorem{COROLLARY}[THEOREM]{Corollary}
\newenvironment{corollary}{\begin{COROLLARY} \hspace{-.85em} {\bf :} }%
                          {\end{COROLLARY}}
\newtheorem{PROPOSITION}[THEOREM]{Proposition}
\newenvironment{proposition}{\begin{PROPOSITION} \hspace{-.85em} {\bf :} }%
                            {\end{PROPOSITION}}
\newtheorem{DEFINITION}[THEOREM]{Definition}
                            {\end{DEFINITION}}
\newtheorem{CLAIM}[THEOREM]{Claim}
                            {\end{CLAIM}}
\newtheorem{EXAMPLE}[THEOREM]{Example}
                            {\end{EXAMPLE}}
\newtheorem{REMARK}[THEOREM]{Remark}
                            {\end{REMARK}}
							\newtheorem{NOTATION}[THEOREM]{Notation}
							                            {\end{NOTATION}}
\newenvironment{proof}{\noindent {\bf Proof:} \hspace{.677em}}%
                     {}

\newcommand{\bbox}{\vrule height7pt width4pt depth1pt}
\newcommand{\eprf}{\bbox\vspace{0.1in}}
\newcommand{\qed}{\eprf}
\DeclareMathAlphabet{\mathitbf}{OML}{cmm}{b}{it}

\newcommand{\sub}{_}
\def\su{^}
\newcommand{\real}{{\mathbb{R}}}

\newcommand{\xx}{{\+ x}}
\newcommand{\yy}{{\+ y}}

\renewcommand{\L}{{\cal L}}

\renewcommand{\P}{{\cal P}}

\newcommand{\set}[1]{\left\{ #1 \right\}}

\newcommand{\eg}{\emph{e.g.}~}
\newcommand{\ie}{\emph{i.e.}~}

\newcommand{\cf}{\emph{cf.}~}

\newcommand{\true}{\mbox{{\sc true}}}

\newcommand{\blemma}{\begin{lemma}}
\newcommand{\elemma}{\end{lemma}}

\newcommand{\bthm}{\begin{theorem}}
\newcommand{\ethm}{\end{theorem}}
\newcommand{\bproof}{\begin{proof}}
\newcommand{\eproof}{\end{proof}}
\newcommand{\bprop}{\begin{proposition}}
\newcommand{\eprop}{\end{proposition}}

\newcommand{\bi}{\begin{itemize}}
\newcommand{\ei}{\end{itemize}}
\newcommand{\be}{\begin{enumerate}}
\newcommand{\ee}{\end{enumerate}}
\newcommand{\beq}{\begin{equation}}
\newcommand{\eeq}{\end{equation}}
\newcommand{\bcase}{\begin{cases}}
\newcommand{\ecase}{\end{cases}}
\newcommand{\diag}{\operatorname{diag}}
\newcommand{\+}[1]{{#1}}%{\ensuremath{\mathbf{#1}}}

\title{The Symbolic Interior Point Method}

\author{
   Martin Mladenov \\
  TU Dortmund Univeristy \\
 \texttt{martin.mladenov@cs.tu-dortmund.de} \\
   \And
  Vaishak Belle\\
  KU Leuven\\
   \texttt{vaishak@cs.kuleuven.be} \\
   \And
  Kristian Kersting \\
  TU Dortmund University \\
  \texttt{kristian.kersting@cs.tu-dortmund.de} \\
}

\begin{document}

\maketitle

\begin{abstract} \small 
  %Numerical optimization is arguably the most prominent computational framework in machine learning and AI, serving as an assembly language for diverse problems such as classification, entropy minimization and computing the optimal policy in sequential decision making. 
%An emerging 
A recent trend in probabilistic inference emphasizes the codification of models in a formal syntax, with suitable high-level features such as individuals, relations, and connectives, enabling descriptive clarity, succinctness and circumventing the need for the modeler to engineer a custom solver. Unfortunately, bringing these linguistic and pragmatic benefits to numerical optimization has proven surprisingly
challenging. In this paper, we turn to these challenges: we introduce a rich modeling language, for which an interior-point method computes approximate solutions in a generic way. While logical features easily complicates the underlying model, often yielding intricate dependencies, we exploit and cache local structure using algebraic decision diagrams (ADDs). 
%It is known that 
Indeed, standard matrix-vector algebra is efficiently realizable in ADDs, but we argue and show that well-known second-order methods are not ideal for ADDs. Our engine, therefore, %uses 
invokes a sophisticated matrix-free approach. We demonstrate the flexibility of the resulting symbolic-numeric optimizer on decision making and compressed sensing tasks with millions of non-zero entries. 

\end{abstract}

%!TEX root = nips_2016.tex

\section{Introduction} % (fold)
\label{sec:introduction}

Numerical optimization is arguably the most prominent computational framework in machine learning and AI. It can seen as an \emph{assembly language} for hard combinatorial problems ranging from classification and regression in learning, to computing optimal policies and equilibria in decision theory, to entropy minimization in information sciences \cite{DBLP:journals/jmlr/BennettP06,puterman,murphy2012machine}. What makes optimization particularly ubiquitous is that, similar to probabilistic inference, it provides a formal apparatus to reason (\ie express preferences, costs, utilities) about  possible worlds. However, it is also widely acknowledged that many AI models are often best described using a combination of natural and mathematical language and, in turn, require  algorithms to be individually engineered. 

An emerging discipline in probabilistic inference  emphasizes the codification of models in a suitable formal syntax \cite{DBLP:conf/ijcai/MilchMRSOK05,richardson2006markov,McAllesterEtAl2008-CSAIL}, enabling the rapid prototyping and solving of complex probabilistic structures. They are driven by the following desiderata: \begin{enumerate}
	\item[(D1)] The syntax should be expressive, allowing high-level (logical) features such as individuals, relations, functions, connectives, in service of  descriptive clarity and succinct characterizations, including context-specific dependencies (\eg pixels probabilistically depend only on neighboring pixels, the flow of quantities at a node is limited to connected nodes). 

	\item[(D2)] The inference engine should be {\it generic,} circumventing the need for the modeler to develop a custom solver. 
\end{enumerate}

Unfortunately, bringing these linguistic and pragmatic benefits to optimization has proven surprisingly challenging. Mathematical programs are specified using linear arithmetic, matrix and tensor algebra, multivariate functions and their compositions. Classes of programs (\eg linear, geometric) have distinct  standard forms, and solvers require models to be painstakingly reduced to the standard form; otherwise, a custom solver has to engineered. The approach taken in recent influential proposals, such as disciplined programming \cite{gb08}, is to carefully constrain the specification language so as to provide a structured interface between the model and the solver, by means of which geometric properties such as the curvature of the objective can be inferred. Basically, when contrasted with the above desiderata, solver genericity is addressed only partially, as it requires the modeler to be well-versed in matrix-vector algebra, and little can be said about high-level features, \eg for codifying dependencies.

In this paper, we are %interested in 
addressing the above desiderata. Doing so has the potential %We believe progress in this direction would 
to greatly simplify the specification and prototyping of AI and ML models \cite{DBLP:journals/cacm/Domingos12,DBLP:journals/jmlr/BennettP06}. 

Let us elaborate on the desiderata in our context. 
The introduction of logical features, while attractive from a modeling viewpoint, assuredly complicates the underlying model, yielding, for example, intricate dependencies. Local structure, on the other hand, can be exploited \cite{Friesen:2015aa}. In probabilistic inference, local structure has enabled tractability in high-treewidth models, culminating in the promising direction %on the use of 
of using circuits and decision-diagrams for the underlying %Bayesian or Markov network 
graphical model~\cite{DBLP:journals/ai/ChaviraD08}. Such data structures are a more powerful representation, in admitting compositionality and refinement. Here, we would like to consider the idea of using such a data structure for optimization, for which we turn to early pioneering research on {\it algebraic decision diagrams} (ADDs) that supports efficient matrix manipulations (compositionality) and caching of submatrices (repeated local structure) \cite{Fujita1997,Clarke1996}. 

Now, suppose a linear program encoded in a high-level language has been to reduced to an efficiently manipulable data structure such as an ADD. %Nonetheless, 
Unfortunately, it is far from obvious how a generic solver can be engineered for it. Matrix operations with ADDs, for example, are efficient only under certain conditions, such as: a) they have to be done recursively, in a specific descent order; b) they have to involve the entire matrix (batch mode), \ie access to arbitrary submatrices is not efficient. This places rather specific constraints on the kind of method that could benefit from an ADD representation, ruling out approaches like random coordinate descent. In this paper, we aim to engineer and  construct a solver for such matrices in circuit representation. We employ ideas from the matrix-free interior point method \cite{DBLP:journals/coap/Gondzio12}, which appeals to an iterative linear equation solver together with the log-barrier method, to achieve a regime where the constraint matrix is only accessed through matrix-vector multiplications.  Specifically, we show %, in particular, 
a  ADD-based realization of the approach leverages the desirable  properties of these representations (\eg caching of submatrices), leading to a robust and fast engine. We demonstrate this claim empirically.

\section{Lineage}
%\section{Prior Art}
\label{related}

Expressiveness in modeling languages for numerical optimization is a focus of many proposals, \eg \cite{fourer1987ampl,wallace2005}. However, they blur the border between declarative and imperative paradigms,  using sets of objects to index 
LP variables and do not embody logical reasoning. Disciplined programming \cite{gb08} enables an object-oriented approach to constructing optimization problems, but falls short of our desiderata as argued before. Taking our cue from statistical relational learning \cite{domingos2006unifying,roth2007}, we argue for algebraic modeling that is fully integrated with classical logical machinery (and not just logic programming \cite{klabjan2009algebraic}). This allows the specification of correlations, groups
 and properties in a natural manner, as also observed elsewhere \cite{kersting2015,gordonHD09}. 

 The efficiency of ADDs for matrix-vector algebra was established in \cite{Clarke1996}. In particular, the use of ADDs for compactly specifying  (and solving) Markov decision processes (\ie representing transitions and rewards as Boolean functions) 
 was popularized in \cite{hoey1999spudd}; see \cite{DBLP:conf/ijcai/ZamaniSDB13,Cui:2016aa} for  recent offerings.  We differ fundamentally from these strands of work in that we are advocating the realization of iterative methods using ADDs, which (surprisingly) has never been studied in great detail to the best of our knowledge. Therefore, we call our approach {\it symbolic numerical optimization}.

%!TEX root = nips_2016.tex

%\renewcommand{\overline}{\bar}

%\section{Logical Preliminaries} % (fold)
\section{Primer on Logic and Decision Diagrams} % (fold)
\label{sec:preliminaries}

We cover some of the logical preliminaries in this section.
%In this section, we introduce the background to our framework, including the logical notions, ADDs, and the syntax for first-order linear constraints. 
% \subsection{Logical Background} % (fold)
% \label{sub:logical_background}
%{\bf Logical Notions} 
To prepare for the syntax of our high-level mathematical programming language, we recap basic notions from mathematical logic \cite{enderton1972mathematical}. A propositional language \( \L \) consists of finitely many propositions \( {\cal P} = \set{p,\ldots,q} \),  from which formulas are built using  connectives \( \set{\overline{~\cdot~}, \lor,\land}. \) A \( \L \)-model \( M \) is a \( \set{0,1} \)-assignment to the symbols in \( \P \), which is extended to complex formulas inductively. For example, if \( \P = \set{p,q} \) and \( M = \langle p = 1, q = 0 \rangle, \) we have \( M \models p \), \( M \models p\lor q, \) \( M\models \overline q \) but \( M \not\models p \land q. \)

% A \( \set{0,1} \)-assignment to \( {\cal P}' \subset {\cal P} \) is referred to as a {\it partial} model. 

The logical language of finite-domain function-free first-order logic  consists of finitely many predicate symbols \( \set{P(x), \ldots, Q(x,y), \ldots, R(x,y,z), \ldots} \) and a domain \( D \) of constants. Atoms are obtained by substituting variables in predicates with constants from \( D \), \eg \( P(a), Q(a,b), R(a,b,c) \) and so on wrt \( D = \set{a,b,\ldots,c}. \) Formulas are built as usual using connectives \( \set{\overline{~\cdot~},  \lor,\land, \forall, \exists} \). Models are \( \set{0,1} \)-assignments to atoms, but can also interpret quantified formulas, \eg \( \forall x[P(x)]. \) For example, if \( D = \set{0,1} \) and \( M = \langle P(0) = 1, P(1) = 1 \rangle \), then \( M \models P(0), M\models P(0)\land P(1) \) and \( M \models \forall x P(x). \) Finally, although finite-domain function-free first-order logic is essentially propositional, it can nonetheless serve as a convinient template for specifying correlations, groups and properties, \cf statistical relational learning \cite{domingos2006unifying}. 

% When modeling the optimization constraints for a domain, we admit the  specification of high-level features over logical notions, such as individuals,  relations and connectives: that is, formulas in a logical system. In general, given a logic \( \L \), formulas \( \F, \) and models \( \M, \) for any \( M\in \M \) and \( \phi\in \F,  \) we write \( M \models \phi \) to mean that \( \phi \) is \textit{true} at \( M. \) 
% %We write \( \models \phi \) to mean that for every \( M\in \M, \) \( M\models \phi. \) 
% For example, in propositional logic over   propositions \( \set{p,q} \), given a model  \( M = \langle p = 1, q = 1 \rangle, \) we have \( M \models p \), \( M \models p\lor q, \) \( M\models p\land q \) but \( M \not\models \overline p. \) Equivalently, a formula can be seen as a Boolean function. 

% and the usual connectives \( \set{\lor,\land,\overline{\cdot}} \), we obtain formulas such as \( \overline p, p\lor q, p\land q \). Models, then,  are \( \set{0,1} \) assignments to propositions. For example, given \( M = \langle p = 1, q = 1 \rangle, \) we have \( M \models p \), \( M \models p\lor q, \) \( M\models p\land q \) but \( M \not\models \overline p. \) Finally, \( \models p \lor \overline p. \)

% subsection logical_background (end)
%
% \subsection{BDDs and ADDs} % (fold)
% \label{sub:bdd_and_add}

%{\bf Decision Diagrams.} 

A BDD \cite{bryant1986graph} is a compact and efficiently manipulable data structure for a Boolean function \( f\colon \set{0,1}\su n \rightarrow \set{0,1}. \) Its roots are obtained by the Shannon expansion of the {\it cofactors} of the function: if \( f\sub x \) and \( f\sub {\overline x} \) denote the partial evaluation of \( f( x, \ldots) \) by setting the variable \( x \)  to \( 1 \) and \( 0 \) respectively, then \begin{equation*}\label{eq:shannon}
	f = x\cdot f \sub x + \overline x \cdot f \sub {\overline x}.
\end{equation*} When the Shannon  expansion is carried out recursively, we obtain a full binary tree whose non-terminal nodes, labeled by variables \( \set{\ldots,x\sub i,\ldots} \), represent a function: its left child is \( f \)'s cofactor w.r.t. \( x\sub i \) for some \( i \) and its right child is \( f\)'s cofactor w.r.t. \( \overline {x\sub i}. \) The terminal node, then, is labeled \( 0 \)  or \( 1 \) and corresponds to a total evaluation of \( f. \) By further ordering the variables, a graph, which we call the (ordered) BDD of \( f, \) can be constructed such that at the {\it k}th level of the tree, the cofactors wrt the {\it k}th variable  are taken. Given an ordering, BDD representations are {\it canonical}: for any two functions \( f,g \colon \set{0,1}\su n \rightarrow \set{0,1} \), \( f \equiv g \Leftrightarrow f\sub x \equiv g\sub x  \) and \( f\sub {\overline x} \equiv g\sub {\overline x} \); and  {\it compact} for Binary operators  \( \circ\in \set{+, \times, \ldots} \): \( |f\circ g| \leq |f||g|. \) 

ADDs generalize BDDs in representing functions of the form \( \set{0,1}\su n \rightarrow \real, \) and so inherit the same structural properties as BDDs except, of course, that terminal nodes are  labeled with real numbers \cite{Fujita1997,Clarke1996}. Consider any real-valued vector of length \( m \): the vector is indexed by \( \lg m \) bits, and so a function of the form \( \set{0,1}\su {\lg m} \rightarrow \real \) maps the vector's indices to its range. Thus, an ADD can represent the vector. By extension, any real-valued  \( 2\su m \times 2 \su n \) matrix \( A \) with row index bits \( \set{x\sub 1, \ldots, x\sub m} \) and column index bits \( \set{y\sub 1, \ldots, y\sub n} \) can be represented as a function \( f(x\sub 1, y\sub 1, x\sub 2, \ldots) \) such that its cofactors are the entries of the matrix. The intuition is to treat \( x\sub 1 \) as the most significant bit, and \( x\sub m \) as the least. Then \( A \) represented by a function \( f \) as an ADD is: \( (f\sub {\overline {x\sub 1} \overline {y\sub 1}} ~~~f\sub {\overline {x\sub 1} {y\sub 1}}) \) as the first row and \( (f\sub {x\sub 1 \overline {y\sub 1}} ~~~f\sub {x\sub 1 y\sub 1}) \) as the second, where each submatrix is similarly defined wrt the next significant bit. Analogously, when multiplying two \( m \)-length vectors represented as Boolean functions \( f,g\colon\set{0,1}\su {\lg m} \rightarrow \real, \) we can write \( [f\sub {\overline {x\sub 1}}~~~f\sub {x\sub 1}] [g\sub {\overline {x\sub 1}}~~~g\sub {x\sub 1}]\su T \), taking, as usual, \( x\sub 1 \) as the most significant bit. 

\begin{wrapfigure}{l}{0.4\textwidth}
  \begin{center}
    \includegraphics[width=0.38\textwidth]{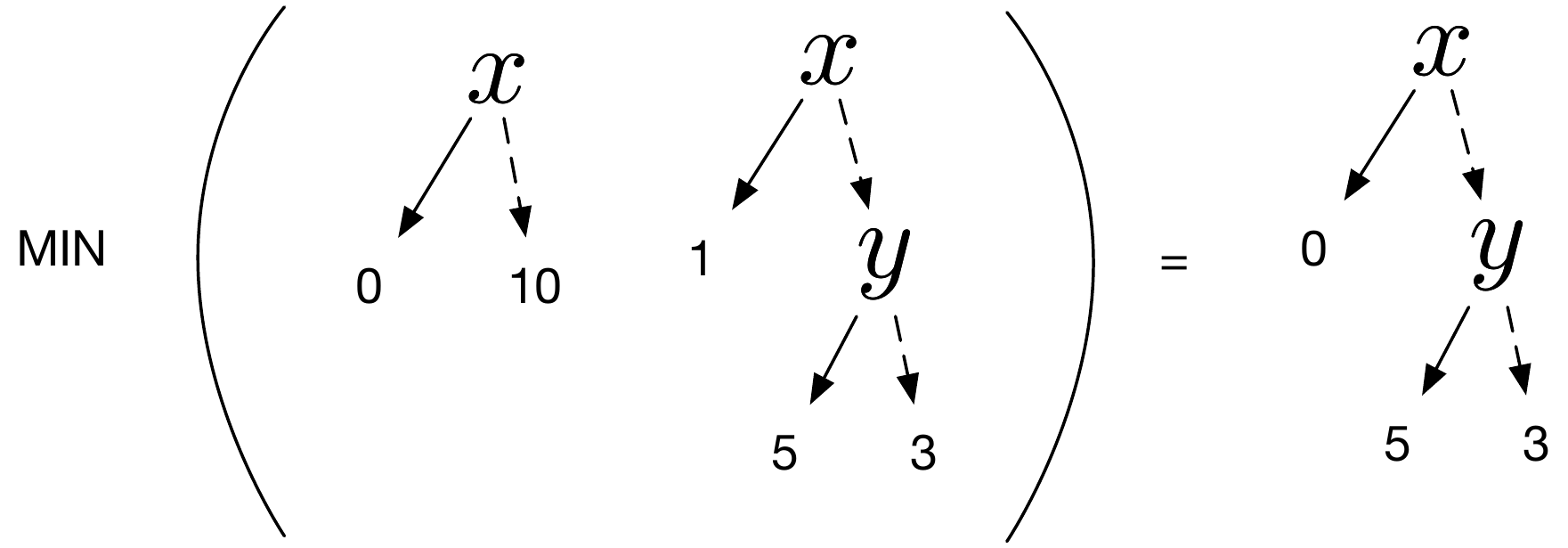}\\[2ex]
    \includegraphics[width=0.25\textwidth]{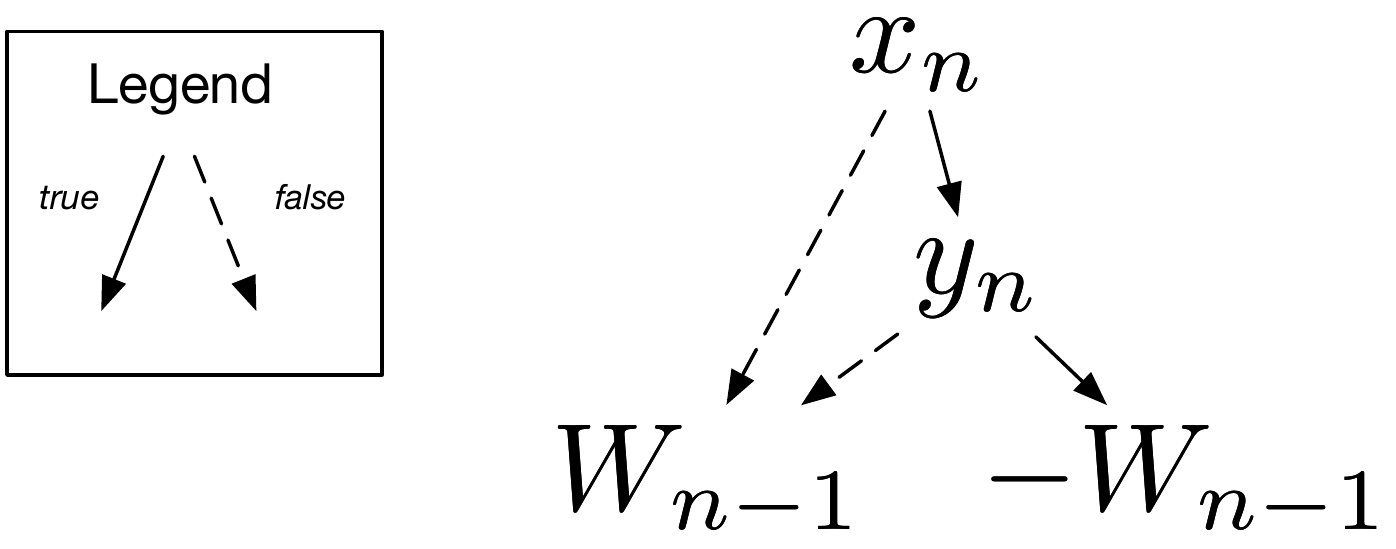}
  \end{center}
  \caption{Minimization of two functions using ADDs (above), and the ADD for the Walsh matrix (below).}\label{fig:add}
\end{wrapfigure}
As can be gathered by the cofactor formulations, the ADD representation admits two significant properties. First, when identical submatrices occur in various parts of the matrix, \eg in block and sparse matrices, different cofactors map to identical functions leading to compact ADDs: a prime example is the Walsh matrix recursively defined as \( W\sub 0 = 1 \), and \( W\sub n \) has \( (W\sub {n-1}~~~W\sub {n-1}) \) as the first row, and \( (W\sub {n-1}~~~-W\sub {n-1}) \) as the second, whose ADD is given in Figure \ref{fig:add}. 
As these functions are cached, significant savings are possible.
% in recursively defined operations for submatrices. that maintain the variable ordering of the ADD. 
%Second, the ADD representation of a matrix of dimension (= number of rows) \( n \) with \( m \ll n \) nonzero entries has space complexity \( O(m\cdot \log n) \), which is superior to standard sparse-matrix representations. Finally, 
Second, matrix algorithms involving recursive-descent procedures on submatrices (\eg multiplication) and  term operations (addition, maximization) are efficiently implementable by manipulating the ADD representation;  see Figure \ref{fig:add}. 
In sum,  standard matrix-vector algebra can be implemented over ADDs efficiently, and in particular, the caching of submatrices in recursively defined operations (while implicitly respecting the variable ordering of the ADD) will best exploit the superiority of the ADD representation.

\section{First-Order Logical Quadratic Programming} % (fold)
\label{sub:first_order_mathematical_programs}

A convex quadratic program (QP) is an optimization problem over the space $\real^n$, that is, we want to to find a real-valued vector $\+x \in \real^n$ from the solution set of a system of linear inequalities $
\{\+x \geq \+0 : \+A\+x = \+b\}$ such that a convex quadratic objective function $f(\+x) = \+x^TQ\+x + \+c^T\+x$ is minimized. For each QP, there exists a complementary (dual) QP-D such that each solution of QP-D provides an upper bound on the minimum of QP, for the maximizer of QP-D, this bound is tight. In this paper we assume that QP and QP-D can be reduced to the following standard form
\begin{tabular}{p{7cm}p{7cm}}
{\begin{align*}
& \text{PRIMAL-QP}:\\
& {
\operatorname{minimize}}
& & c^Tx + 1/2x^TQx \\
& \text{subject to}
& & Ax = b, \\
& & & x \geq 0.
\end{align*}
}
&
{\begin{align*}
& \text{DUAL-QP}:\\
&{\operatorname{maximize}}
& & b^Ty - 1/2x^TQx \\
& \text{subject to}
& & A^Ty + s - Qx = c , \\
&&& s \geq 0.
\end{align*} }
\end{tabular}
Here, $Q$ is positive semi-definite. Whenever $Q = 0$, we speak of linear programs (LPs).
% A linear program (LP) is an optimization We take this specification as the {\it canonical} (ground) form for the linear program and use the notation $G = (\+A,\+b,\+c)$ for specifying LPs.  Solutions to linear programs can be approximated efficiently  by algorithms such as the log-barrier method \cite{papadimitriou1982combinatorial}. 

% which we will discuss later on. 
%Informed by advances in probabilistic inference and declarative machine learning (SRL; Domingos; Church), 

In this paper, we would like to provide an expressive modeling language with high-level (logical) features such as individuals, relations, functions, and connectives. While (some) high-level features are prominent in many optimization packages such as AMPL \cite{opac-b1123349}, they are reduced (naively) to canonical forms transparently to the user, and do not embody logical reasoning. Logic programming is supported in other proposals \cite{klabjan2009algebraic}, but their restricted use of negation makes it difficult to understand the implications when modeling a domain. We take our cue from statistical relational learning \cite{domingos2006unifying}, as considered in \cite{kersting2015}, to support any (finite) fragment of first-order logic with classical interpretations for operators. Given a logical language \( \L, \)
the syntax for first-order logical quadratic programs is:
% There is a growing interest in expressing LPs using high-level features (AMPL,Boyd). In this work, we advocate the use of a compact representation that generates linear constraints by appealing to  a logical system, such as propositional logic or (finite domain) first-order logic. 
% In particular, 
% %
% % A more compact representation can be obtained by appealing to a logical system, such as propositional logic or (finite domain) first-order logic. For our purposes,
% we 
% define a first-order linear constraint using the notation: \[
% 	\forall \+y.~\psi(\+y): \forall \+x.~\sum\sub {\set{\+x: \phi(\+x, \+y)}}  a(\+x,\+y)v(\+x) \geq b(\+y) 
% \]
% where, we write \( \delta(\+z) \) to mean that the variables \( \+z \) are free in \( \delta \). The expression is to be read procedurally, and says: for every possible assignment to \( \+y, \) for every model \( M \) of \( \psi(\+ y) \), consider constraints \( a(\+x,\+y)v(\+x) \geq b(\+y)  \) generated by every possible assignment to \( \+x \) satisfying \( \phi(\+x,\+y) \). 
% In particular, 
% we 
% define a first-order linear constraint using the notation:
 \[
\underset{v}{\operatorname{min.}} \sum_{\set{\+x,\+x^\prime: \varphi(\+x,\+x^\prime)} } \!\!\!\!\! q(\+x,\+x^\prime)  v(\+x) v(\+x^\prime)+ \sum_{\set{\+x: \psi(\+x)} } \!\!\!\!\! c(\+x)  v(\+x) ~~~
\text{s.t.}~~~	\set{\+y: \psi(\+y)}: \sum\sub {\set{\+x: \phi(\+x, \+y)}}  a(\+x,\+y)v(\+x) \geq b(\+y) 
\]
where, we write \( \delta(\+z) \) to mean that the \( \L \)-formula \( \delta \) mentions 
the logical symbols \( \+z \), and read \( \set{\+z: \delta(\+z)} \) as the set of all assignments to \( \+z \) satisfying \( \delta(\+z). \) The constraints are to be read procedurally as follows: for every assignment to \( \+y \) satisfying \( \psi(\+ y) \),
%(partial) model \( M \) of \( \psi(\+ y) \), 
consider the constraint \( \sum\sub C a(\+x,\+y)v(\+x) \geq b(\+y)  \), where the set \( C \) are those   assignments to \( \+x\) satisfying \( \phi(\+x,\+y) \). For example, consider: \begin{align*}
\operatorname{minimize}_{v}  \quad& \sum_{\set{x: \true} } \!\!\!\!\! v(x) ~~~
\text{subject to} & 
&   \set{y: \true}: \sum_{\set{x\colon x\vee y}} v(x) \geq 1, 
&  \set{y: \true}: v(y) \geq 0\;.
\end{align*}
%Here, \( y\lor \neg y \) is tautologous, of course, but they serve the purpose of explicitly mentioning the variables that are considered in the partial models. Basically, 
That is, we are to minimize \( v(x) + v(\overline x) \) subject to \( v(x) \geq 1 \) and \( v(\overline x) \geq 1 \) (for the cases where \(\overline y \)), \( v(x) + v(\overline x) \geq 1 \) and \( v(x) \geq 0 \) (for the cases where \( y \)), yielding the following canonical form: \[\+A =  \begin{bmatrix}
 0~ (\overline x \overline y) & 1~ (x \overline y)\\
1~ (\overline x y) & 1~ (x y)\\
 1~ (\overline x \overline y) & 0~ (x \overline y) \\
 0~ (\overline x y) & 1~ (x y) \\
 \end{bmatrix},~~ 
 \quad \+b = \begin{bmatrix}
1\\
1\\
0\\
0\\
 \end{bmatrix}, \quad \+c = \begin{bmatrix}
1\\
1\\
 \end{bmatrix}. 
 \]
By extension, an example for a relational constraint is one of the form: \( \sum\sub{x:\exists z~ \textrm{\it Friends}(z,x)} v(x) \). Here, wrt \( D = \set{a,b,c} \) and a database \( \set{{\it Friends}(b,a), {\it Friends}(b,c)} \), we would instantiate the constraint as \( v(b) + v(c). \) For simplicity, however, we limit discussions to propositions in the sequel. 

One of the key contributions in our engine is that we are able to transform the constraints in such a high-level mathematical program directly to an ADD representation. (Quantifiers are eliminated as usual: existentials as disjunctions and universals as conjunctions \cite{enderton1972mathematical}.) The rough idea is this. Order the variables \( \+y \) in \( \psi(\+y). \) Any propositional formula can be seen as a Boolean function mapping to \( \set{0,1}. \) That is, assignments to \( \yy \) determine the value of the corresponding Boolean function, \eg \( \langle y\sub 1=1, y\sub 2 = 0\rangle \) maps the function \( y\sub 1 \land y\sub 2 \) to 0. We build the ADD for \( \psi(\yy) \), and for each complete  evaluation of this function, we build the ADD for \( \phi(\xx,\yy) \) and let its leaf nodes be mapped to \( b(\yy) \). We omit a full description of this procedure, but go over the main result:

\begin{theorem} There is an algorithm for building an ADD for terms from a first-order logical mathematical program without converting the program to the canonical (ground) form. 
The ADD obtained from the algorithm is identical to the one obtained from the ground form. 
%, on the one hand, and the ADD 
%representation for the ground form, on the other, are identical. 
 \end{theorem} 

 \begin{proof} We simply need to argue that a term of a first-order logical mathematical program corresponds to a Boolean function \( \set{0,1}\su n \rightarrow \real \), where \( n \) is the number of propositions. The procedure described above builds its ADD. By the canonicity of ADDs, the claim follows. \qed 
 	
 \end{proof}

%{\bf correct description todo}

%To see an example, consider the following first-order linear constraint {\bf example from notes}.

%!TEX root = nips_2016.tex

\section{Solution Strategies for First-Order Logical QPs}

Given the representation language, we now turn towards
solving logical QPs.
To prepare for discussions, % on solution strategies, 
let us establish an elementary notion of algorithmic correctness for ADD implementations. We assume that given a first-order logical mathematical program, we have in hand the ADDs for \( A, b \) and \( c. \) Then, it can be shown: 

\begin{theorem} Suppose \( A, b \) and \( c \) are as above, and \( e \) is any arithmetic expression over them involving standard matrix binary operators. Then there is a sequence of operations over their ADDs, yielding a function \( h, \) such that \( h = e. \)
\end{theorem} 

The kinds of expressions we have in mind are \( e = Ax - b \) (which corresponds to the residual in the corresponding system of linear equations). 
%The proof is straightforward: 
The proof is as follows:

\begin{proof} For any \( f,g: \set{0,1}\su n \rightarrow \real, \)  and (standard) binary matrix operators (multiplication, summation, subtraction), observe that  \( h = f\circ g \) iff \( h\sub x = f\sub x \circ g\sub x \) and \( h\sub {\overline x} = f\sub{\overline x} \circ g\sub {\overline x} \). 
By canonicity, the ADD for \( h \) is precisely the same as the one for \( f  \) and \( g \) composed over \( \circ. \) In other words, for any arithmetic expression \( e \) over \( \set{f,g,\circ} \), the ADD realization for $h$  is precisely the same function. \qed 
  
\end{proof} 

To guide the construction of the engine, we will briefly go over the operations previously established as efficient with ADDs \cite{Fujita1997,Clarke1996}, and some implications thereof for a solver strategy. 

\begin{theorem}\cite{Fujita1997,Clarke1996}  Suppose \( A, A' \) are real-valued  matrices. Then the following can be efficiently implemented in ADDs using recursive-decent procedures: 1) accessing and setting a submatrix $A^ *$ of $A$; 2) termwise operations, \ie \( (A\circ A')\sub {ij} = A\sub {ij} \circ A' \sub {ij} \) for any termwise operator \( \circ \); and 3) vector and matrix multiplications.
  
\end{theorem} 

The proof for these claims and the ones in the corollary below always proceed by leveraging the Shannon expansion for the corresponding Boolean functions, as shown in Section \ref{sec:preliminaries}. We refer interested readers to \cite{Fujita1997,Clarke1996} for  the complexity-theoretic properties of these operations. It is worth noting, for example, that multiplication procedures that perform both  (naive) block computations and ones based on Strassen products can be recursively defined. For our purposes, we get: 

\begin{corollary}\label{cor:vectorfast} Suppose $\+d = [d\sub 1~~\cdots~~d\sub m]$ and $\+e$ are $m$-length real-valued vectors, and $k$ is a scalar quantity. Then the following can be efficiently implemented in ADDs  using recursive-descent procedures: 1) scalar multiplication, \eg $k\cdot \+d$; 2) vector arithmetic, \eg $\+d + \+e$; 3) element sum, \eg $\sum\sub i d\sub i$; 4) element function application, \eg if $w\colon \real\rightarrow \real$, then computing $w(d) =  [w(d\sub 1)~~\cdots~~w(d\sub m)]$; and 5) norms, \eg $\|d\|$ and $\|d\|^2$.
\end{corollary}

%\subsection{The (Naive) Ground Approach}
\subsection{A Naive Ground-and-Solve Method}
The most straightforward approach for solving a first-order logical QP (FOQP) is to reduce the problem to the normal form and use a standard solver for QPs, such as an interior point method, an augmented Lagrangian method, or some form of active set method, \eg generalized simplex. The correctness of this solution strategy is guaranteed by the semantics of the first-order constraints: in general, every FOQP can be brought to the normal form. While this method would work, it exhibits a significant drawback in that the optimization engine cannot leverage knowledge about the symbolic structure of the problem. That is, even if the problem compiles to a very small ADD, the running time of the optimizer will depend (linearly at best) on the number of nonzeros in the ground matrix. Clearly, large dense problems will be completely intractable. However, as we will see later, some dense problems can still be attacked with the help of structure.
% * <vaishak@me.com> 2016-05-20T12:29:03.497Z:
%
% > eqref
%
% needs to be resolved
%
% ^.

\subsection{The Symbolic Interior Point Method}

While the ground-and-solve method is indeed correct, one can do significantly better, as we will now show. In this section, we will construct a solver that automatically exploits the symbolic structure of the FOQP, in essence, by appealing to the strengths of the ADD representation. The reader should note that much of this discussion is predicated on problems having considerable logical structure, as is often the case in real-word problems involving  relations and properties. %enabled in many real-world problems with relations and properties.) 

Recall from the previous discussion that in the presence of cache,  (compact)  ADDs translate to very fast matrix-vector multiplications. Moreover, from Corollary \ref{cor:vectorfast}, vector operations are efficiently implementable, which implies that given an ADD for $A$,  $\+x$ and $\+b$, the computation of the residual $\+y = A\+x -\+b$ is more efficient than its matrix counterpart \cite{Fujita1997}. Analogously, a descent along a direction ($\+x_k = \+x_{k-1} + \alpha \Delta\+x$) given ADDs for $\+x_{k-1}$ and $\Delta x$ has the same run-time complexity as when performed on dense vectors. 

We remark that for ADD-based procedures to be efficient, we need to respect the variable ordering implicit in the ADD. Therefore, the solver strategy rests on the following constraints:  i) {\it the engine must manipulate  the matrix only through recursive-descent arithmetical operations, such as matrix-vector multiplications (matvecs, for short);} and ii) {\it operations must manipulate either the entire matrix or those submatrices corresponding to  cofactors (\ie arbitrary submatrices are non-trivial to access).} 

%Note, however, that in order for the ADD matvec to be efficient, it needs to be done in the recursion order defined by the ADD and has to be done on the entire matrix. Hence, when constructing a solver, we need to satisfy the following constraints as well as possible: a) the solver must access $A$ only through matvecs, and access the current solution only through vector arithmetic and seperable functions; b) all updates must be done in bulk (using the entire $A$ and $\+x$), i.e. we need to avoid accessing random submatrices of $A$ as much as we can.   

We will now investigate a method that satisfies these requirements.
We proceed in two steps. In step $1$, we will demonstrate that solving a QP can be reduced to solving a sequence of linear equations over $A$. Next, in step $2$, we will make use of an iterative solver that computes numerical solutions by a sequence of residuals and vector algebraic operations. As a result, we will obtain a method that fully utilizes the strengths of ADDs. Due to space constraints, we will not be able to discuss the construction in full detail, and so we sketch the main ideas that convey how ADDs are exploited. 
%and convince the readers that the resulting approach can make full use of the benefits provided by ADDs. 
%todo \textbf{To fill in the gaps, we provide a full reference implementation, including a symbolic environment to specify FOQPs, a compiler from FOQP to ADD, and an ADD numerical optimizer. This implementation can be found at (TODO: where?). }
% * <vaishak@me.com> 2016-05-20T13:14:39.352Z:
%
% > To fill in the gaps, we provide a full reference implementation, including a symbolic environment to specify FOQPs, a compiler from FOQP to ADD, and an ADD numerical optimizer. This implementation can be found at (TODO: where?). 
%
% hmm. how do we do this anon submision 
%
% ^.
{\footnotesize \begin{figure}
\begin{subfigure}{.45\textwidth}
    \begin{minipage}{\linewidth}

\begin{algorithm}[H]
\tiny  
%\linesnumbered
%\SetAlgoLined
\SetKwFunction{newColor}{newColor}
\KwIn{$(x^0, y^0, s^0)$ with $(x^0, s^0) \geq 0$}

$k \leftarrow 0$\;
\While{stopping criterion not fulfilled}{
  Solve \eqref{eq:normeq} with $(x, y, s) = (x^k, y^k, s^k)$ to obtain a direction $(\Delta x^k, \Delta y^k, \Delta s^k )$\;
  Choose step length $\alpha_k \in (0,1]$\;
  Update $(x^{k+1}, y^{k+1}, s^{k+1}) \leftarrow \alpha_k(\Delta x^k, \Delta y^k, \Delta s^k)$
  $k \leftarrow k+1$
}
\Return{$\mathbf{x}^k$}\;

\end{algorithm}
\end{minipage}
\caption{Primal-Dual Barrier Method \label{alg:logbar}}
    \end{subfigure}
    \quad\quad 
    % =====================
    \begin{subfigure}{.45\textwidth}
     \begin{minipage}{\linewidth}
       \begin{algorithm}[H]
\tiny  
\SetKwFunction{newColor}{newColor}
\KwIn{$\mathbf{A} \in \mathbb{R}^{n\times n}$, $\mathbf{b} \in \mathbb{R}^n$}
$k \leftarrow 0,\; r_0 \leftarrow b - Ax_0$\;
\While{stopping criterion not fulfilled}{
  $k \leftarrow k+1$\;
  \If{k = 1} {
    $p_0 \leftarrow r_0$\;
  }
  \Else{
    $\tau_{k-1} = (r^T_{k-1}r_{k-1})/(r^T_{k-2}r^T_{k-2})$\;
    $p_k = r_{k-1} + \tau_{k-1}p_{k-1}$\;
  }
  $\mu_k = (r^T_{k-1}r_{k-1})/(p_k^TAp_k)$\;
  $\+x^{k} = x_{k-1} + \mu_k p_k$ \;
  $\+r^{k} = r_{k-1} - \mu_k A p_k$ \;
}
\Return{$\mathbf{x}^k$}\;
\end{algorithm}
\end{minipage}
\caption{Conjugate Gradient Method \label{alg:cg}}
    \end{subfigure}% need this comment symbol to avoid overfull hbox
\end{figure}
}

{\bf Step 1: From linear programs to linear equations.}  
 A prominent solver for QPs in standard form is the primal-dual barrier method, see \eg \cite{potra00jcam},  sketched in Alg.~\ref{alg:logbar}. This method solves a perturbed version of the first-order necessary conditions (KKT conditions) for QP:
\begin{align*}
Ax = b,~~
-Qx + A^Ty + s = c,~~
XSe = \mu e,~~
(x,s) \geq 0,
\end{align*}
where $X = \diag(x_1,\ldots,x_n)$, $S = \diag(s_1,\ldots,s_n)$, and $\mu \geq 0$. 
%The precise details of the primal-dual method are not relevant for our discussion  \cite{potra00jcam,DBLP:journals/coap/Gondzio12}. 
%hence we will gloss over them in the interest of space (the reader is referred to potra/gondzio for introductions). 
The underlying idea is as follows \cite{potra00jcam,DBLP:journals/coap/Gondzio12}: by applying a perturbed Newton method to the equalities in the above system, the algorithm progresses the current solution along a direction obtained by solving the following linear system: 
\begin{equation}
 \begin{bmatrix}
       A & 0 & 0           \\[0.3em]
       -Q & A^T & I \\[0.3em]
       S & 0 & X
     \end{bmatrix}
     \begin{bmatrix}
       \Delta x \\[0.3em]
        \Delta y \\[0.3em]
        \Delta s
     \end{bmatrix} = 
     \begin{bmatrix}
        b - Ax \\[0.3em]
       c - A^Ty - s \\[0.3em]
        \mu e - Xs
     \end{bmatrix}\;,
     \label{eq:newtdir}
     \end{equation}
where $e$ is a vector of ones. Observe that besides $A$ and $Q$, which we already have decision diagrams for, the only new information that needs to be computed is in the form of residuals in the right-hand side of the equation. By performing two pivots, $\Delta x$ and $\Delta s$ can be eliminated from the system, reducing it to the so-called \textit{normal equation}:
\begin{equation}
\label{eq:normeq}
A(Q + \Theta^{-1})^{-1}A^T\Delta y = f,
\end{equation}
where $\Theta$ is the diagonal matrix $\Theta_{ii} = \frac{x_i}{s_i}$  \cite{potra00jcam,DBLP:journals/coap/Gondzio12}. Once $\Delta y$ is determined, $\Delta x$ and $\Delta s$ are recovered from $\Delta y$ as $\Delta x = (Q + \Theta)^{-1}(A^T\Delta y - g)$ and $\Delta s = h - A^T\Delta y$, where $h$ and $g$ are obtained from the residuals via vector arithmetic. The reader will note that constructing the left-hand side involves the matrix inverse $(Q + \Theta^{-1})^{-1}$. The efficiency of computing this inverse cannot be guaranteed with ADDs, unfortunately; therefore, we assume that either the problem is separable ($Q$ is diagonal), in which case computing this inverse reduces to computing the reciprocals of the diagonal elements (an efficient operation with  ADDs), or that we have only box constraints, meaning that $A$ is a diagonal matrix, in which case
%$A$ is diagonal and 
solving the equation reduces to solving $(Q + \Theta)\Delta y^{\prime} = A^{-1}f$ and re-scaling. (The general case of going beyond these assumptions is omitted here for space reasons  \cite{DBLP:journals/coap/Gondzio12}.)

%, but omitted here for brevity. however, for the sake of brevity we opt to focus on the simpler cases.  
% * <vaishak@me.com> 2016-05-20T13:24:29.183Z:
%
% > The general case can still be realized, however, the discussion becomes significantly more involved, so we opt to focus on the simpler cases.  
%
% don't follow this 
%
% ^ <vaishak@me.com> 2016-05-20T13:26:43.494Z.

The benefit of reformulating \eqref{eq:newtdir} into \eqref{eq:normeq} can be appreciated from the observation that \ref{eq:normeq} becomes positive-semidefinite, which is crucial for solving this system by residuals. 
%This approach also has some drawbacks, which will be discussed later on.
To reiterate: the primal-dual barrier method solves a quadratic program iteratively by solving one linear system \eqref{eq:normeq} in each iteration. Constructing the right-hand side of this linear system only requires the calculation of residuals and vector arithmetic, which can be done efficiently with ADDs. Moreover, this system does not require taking arbitrary submatrices of $A$ or $Q$. Hence, the primal-dual barrier method meets our requirements. Now, let us investigate how \ref{eq:normeq} can be solved via a sequence of residuals.   

{\bf Step 2: From linear equations to residuals.}  To solve \eqref{eq:normeq}, we employ the conjugate gradient method \cite{Golub:1996:MC:248979}, sketched in Fig.~\ref{alg:cg}. Here, the algorithm uses three algebraic operations: 1) matrix-vector products; 2) norm computation ($r^Tr$) and 3) scalar updates. From Corollary \ref{cor:vectorfast}, all of these operations can be implemented efficiently in  ADDs.
There is, however, one challenge that remains to be addressed. As the barrier method approaches the solution of the QP, the iterates $s^k$ and $x^k$ approach complementary slackness ($x_i s_i = 0)$. This means that the diagonal entries of the matrix $\Theta$ in \eqref{eq:normeq} tend to either $0$ or $+\infty$, making the condition number of \eqref{eq:normeq} unbounded. This is a severe problem for any iterative solver, as the number of iterations required to reach a specified tolerance becomes unbounded. To remedy this situation, Gondzio \cite{DBLP:journals/coap/Gondzio12} proposes the following approach: first, the system can be regularized to achieve a condition number bounded by the largest singular value of $A$. Second, due to the IPM's remarkable robustness to inexact search directions, it is not necessary to solve the system completely. In practice, decreasing the residual by a factor of $0.01$ to $0.0001$ has been found sufficient. Finally, a partial pivoted Cholesky factorization can be used to speed-up the convergence. That is, perform a small number $k$ (say $50$) Cholesky pivots, and use the resulting trapezoidal matrix as a preconditioner. More details on this can be found in \cite{DBLP:journals/coap/Gondzio12}. As demonstrated in \cite{DBLP:journals/coap/Gondzio12} this approach does lead to a practical algorithm. Unfortunately, in our setting, constructing this preconditioner requires that we query $k$ rows of $N$ and perform pivots with them. This forces us to partially back down on our requirement (ii), since we need random access to $k$ rows. However, by keeping $k$ small, we can guarantee the ADD in this unfavorable regime will be kept to a minimum. 
% * <vaishak@me.com> 2016-05-20T13:32:34.711Z:
%
% > condition number of \eqref{eq:normeq} unbounded.
%
% does this need more explanation 
%
% ^ <vaishak@me.com> 2016-05-20T13:32:43.752Z.

Thus, we have a method for solving quadratic programs, implemented completely with ADD operations, and much of this work takes full advantage of the ADD representation (thereby, inheriting its superiority). 
%The method satisfies the first two requirements and for the most part the third one. 
Intuitively, one can expect significant speed-ups over matrix-based methods when the ADDs are compact, \eg arising from structured (in a logical sense) problems. 
% * <vaishak@me.com> 2016-05-20T13:45:36.999Z:
%
% ^.

\section{Empirical Illustration}
Here, we aim to investigate the empirical performance of our ADD-based interior point solver. There are three main questions we wish to investigate, namely: {\bf (Q1)} in the presence of symbolic structure, does our ADD-based solver perform better than its matrix-based counterpart? {\bf (Q2)} On structured sparse problems, does solving with ADDs have advantages over solving with sparse matrices? And, {\bf (Q3)}, can the ADD-based method handle dense problems as easily as sparse problems? 

To evaluate the performance of the approach, we implemented the entire pipeline described here, that is, a symbolic environment to specify QPs, a compiler to ADDs, based on the popular CUDD package, and the symbolic interior-point method described in the previous section.   

To address Q1 and Q2, we applied the symbolic IPM on the problem of computing the value function of a family of Markov decision processes used in \cite{hoey1999spudd}. These MDPs concern a factory agent whose task is to paint two objects and connect them. A number of operations (actions) need to be performed on these objects before
painting, each of which requires the use special of tools,
which are possibly available. Painting and connecting can be done in different ways, yielding results of various quality, and each requiring  different tools. The final product is rewarded according whether the required level of quality is achieved. 
Since these MDPs admit compact symbolic representations, we consider them good candidates to illustrate the potential advantages of symbolic optimization. The computation of an MDP value function corresponds to the following LP: $\operatorname{min.} \sum_{\+s : \mathtt{state}(\+s)} v(\+s)\;, \text{s.t.}\; \{ \+s: \mathtt{state}(\+s), \+a:\mathtt{act}(\+a)\}: v(\+s) \geq \mathtt{rew}(\+s) + \gamma \sum_{\+s^\prime : \mathtt{state}(\+s^\prime)}\mathtt{tprob}(\+s, \+a, \+s^\prime)v(\+s^\prime)$, where $\+s, \+s^\prime, \+a$ are vectors of Boolean variables, $\mathtt{state}$ is a Boolean formula whose models are the possible states of the MDP, $\mathtt{act}$ is a formula that models the possible actions, and $\mathtt{rew}$ and $\mathtt{tprob}$ are pseudo-Boolean functions that model the reward and the transition probability from $\+s$ to $\+s^\prime$ under the action $\+a$. We compared our approach to a matrix implementation of the primal-dual barrier method, both algorithms terminate at the same relative residual, $10^{-5}$. The results are summarized in the following table.

\begin{center}
\begin{tabular}{ |l|c|c|c|c|c|c| }
  \hline
  \multicolumn{4}{|c|}{Problem Statistics} & \multicolumn{2}{|c|}{Symbolic IPM} & Ground IPM\\
  name & \#vars & \#constr & $nnz(A)$ & |ADD| & time[s] & time[s]\\
  \hline
  factory & 131.072 & 688.128 & 4.000.000 & 1819 & 6899 & {\bf 516}\\
  factory0 & 524.288 & 2.752.510 & 15.510.000 & 1895 & {\bf 6544} & 7920 \\
  factory1 &  2.097.150 & 11.000.000 & 59.549.700 & 2406 & {\bf 34749} & 159730\\
  factory2 & 4.194.300 & 22.020.100 & 119.099.000 & 2504 & {\bf 36248} & $\geq$ 48hrs. \\
  \hline
\end{tabular}
\end{center}
The symbolic IPM outperforms the matrix-based IPM on the larger instances. However, the most striking observation is that the running time depends mostly on the size of the ADD, which essentially translates to scaling sublinear in the number of nonzeroes in the constraints matrix. To the best of our knowledge, no generic method, sparse or dense, can achieve this scaling behavior. This provides an affirmative answer to Q1 and Q2.

Our second illustration concerns the problem of compressed sensing (CS). That is, we are interested in recovering the sparsest solution to the problem  $||Ax - b||^2$, where $A\in\mathbb{R}^{m\times n}$ and $n >> m$ where $n$ is the signal length and $m$ the number of measurments. While this is a hard combinatorial problem, if the matrix $A$ admits the so-called Restricted Isometry Property, the following convex problem (Basis Pursuit Denoising - BPDN): $\operatorname{min.}_{\+x\in \mathbb{R}^n} \tau ||x||_1 + ||Ax - b||^2_2$ recovers the exact solution. The matrices typically used in CS tend to be dense, yet highly structured, often admitting an $O(n\log n)$ Fast Fourier Transform-style recur-and-cache matrix-vector multiplication scheme, making BPDN solvers efficient. A remarkable insight is that these fast recursive transforms are very similar to the ADD matrix-vector product. In fact, if we take $A$ to be the Walsh matrix $W_{\log(n)}$ which admits an ADD of size $O(\log n)$, the ADD matrix-vector product resembles the Walsh-Hadamard transform. Moreover, this compact ADD is extracted automatically from the symbolic specification of the Walsh matrix.  We illustrate this in the following experiment. We apply the Symbolic IPM  to the BPDN reformulation of \cite{Fountoulakis2014}, with the Walsh matrix specified symbolically, recovering random sparse vectors. We compare to the reference MATLAB code provided by Gondzio et al, which we have modified to use MATLAB's implementation on the fast Walsh-Hadamard transform. The task is to recover a sparse random vector with $k=50$ normally distributed nonzero entries. Results below.

%The results are summarized below. 

\begin{center}
\begin{tabular}{ |c|c|c|c| }
  \hline
  \multicolumn{2}{|c|}{Problem Statistics} & Symbolic IPM & FWHT IPM \cite{Fountoulakis2014} \\
  n  & m  & time[s] & time[s]\\
  \hline
  $2^{12}$ & $2^{10}$ & {\bf 8.3} & 18  \\
  $2^{13}$ & $2^{11}$ & {\bf 20.2} & 28.2\\
  $2^{14}$ & $2^{12}$ & 46.5 & {\bf 43.9} \\
  $2^{15}$ & $2^{13}$ &  99.2 & {\bf 65.7}\\
  \hline
\end{tabular}
\end{center}
While the method does not scale as well as the hand-tailored solution, we demonstrate that the symbolic approach can handle dense matrices reasonably well, supporting the positive answer of Q3.
% subsection first_order_mathematical_programs (end)

\section{Conclusions} % (fold)
\label{sec:conclusions}

A long-standing goal in machine learning and AI, which is also reflected in the philosophy on the {\it democratization of data,} is to make the specification and solving of real-world problems simple and natural, possibly even for non-experts. To this aim, we considered first-order logical mathematical programs that support individuals, relations and connectives, and and developed a new line of research of symbolically solving these programs in a generic way. In our case, a matrix-free interior point method was argued for. Our empirical results demonstrate the flexibility of this research direction. The most interesting avenue for future work is to explore richer modeling languages paired with more powerful circuit representations. 

% section conclusions (end) 

{\bf Acknowledgements} The authors would like to thank the anonymous reviewers for their feedback. The work was partly supported by the 
DFG Collaborative Research Center SFB 876, project A6.

\small 
\bibliographystyle{IEEEtran}
\bibliography{main,biblio}
%\bibliography{/Users/vaishakbelle/Dropbox/Papers/main}

\end{document}